\documentclass[twoside]{article} 

\usepackage{times}
\usepackage{graphicx} % more modern
\usepackage{subfigure} 

\usepackage{natbib}

\usepackage{algorithm}
\usepackage{algorithmic}

\usepackage[draft]{hyperref}

\usepackage[accepted]{icml2017} 

\usepackage[utf8]{inputenc} % allow utf-8 input
\usepackage[T1]{fontenc} %use 8-bit T1 fonts
\usepackage{hyperref} %hyperlinks - AAAI does not support this
\usepackage{url} % simple URL typesetting
\usepackage{booktabs} % professional-quality tables
\usepackage{amsfonts} % blackboard math symbols
\usepackage{nicefrac} % compact symbols for 1/2, etc.
\usepackage{microtype} %microtypography
\usepackage{amsmath}
\usepackage{breqn}

\usepackage{color}
\usepackage[T1]{fontenc}

\usepackage{graphicx}
\graphicspath{{./figs/}}
\usepackage{epstopdf}
\usepackage{ifthen}

\newcommand{\beq}[1][]{\begin{equation}}
\newcommand{\eeq}{\end{equation}}
\newcommand{\beqr}{\begin{eqnarray}}
\newcommand{\eeqr}{\end{eqnarray}}
\newcommand{\beqrn}{\begin{eqnarray*}}
\newcommand{\eeqrn}{\end{eqnarray*}}

\icmltitlerunning{Hierarchical Subtask Discovery With Non-Negative Matrix Factorization}

\begin{document} 

\twocolumn[
\icmltitle{Hierarchical Subtask Discovery With Non-Negative Matrix Factorization}



% you can specify symbols, otherwise they are numbered in order

\begin{icmlauthorlist}
\icmlauthor{Adam C. Earle}{wits}
\icmlauthor{Andrew M. Saxe}{hv}
\icmlauthor{Benjamin Rosman}{wits,csir}
%\icmlauthor{Iaesut Saoeu}{ed}
%\icmlauthor{Fiuea Rrrr}{to}
%\icmlauthor{Tateu H.~Yasehe}{ed,to,goo} 
%\icmlauthor{Aaoeu Iasoh}{goo}
%\icmlauthor{Buiui Eueu}{ed}
%\icmlauthor{Aeuia Zzzz}{ed}
%\icmlauthor{Bieea C.~Yyyy}{to,goo}
%\icmlauthor{Teoau Xxxx}{ed}
%\icmlauthor{Eee Pppp}{ed}
\end{icmlauthorlist}

%\icmlaffiliation{to}{University of Torontoland, Torontoland, Canada}
%\icmlaffiliation{goo}{Googol ShallowMind, New London, Michigan, USA}
%\icmlaffiliation{ed}{University of Edenborrow, Edenborrow, United Kingdom}

\icmlaffiliation{hv}{Center for Brain Science, Harvard University}
\icmlaffiliation{wits}{School of Computer Science and Applied Mathematics, University of the Witwatersrand}
\icmlaffiliation{csir}{Council for Scientific and Industrial Research, South Africa}

%\icmlcorrespondingauthor{Cieua Vvvvv}{c.vvvvv@googol.com}
%\icmlcorrespondingauthor{Eee Pppp}{ep@eden.co.uk}

\icmlcorrespondingauthor{Adam Earle}{adam.earle@students.wits.ac.za}

\icmlkeywords{boring formatting information, machine learning, ICML}

\vskip 0.3in
]



\printAffiliationsAndNotice{\icmlEqualContribution} % otherwise use the standard text.
%\footnotetext{hi}

%%%%%%%%%%%%%%%%%%%%%%%%%%%%%%%%%%%%%%%%%%%%%%
%%%% ICML style ending
%%%%%%%%%%%%%%%%%%%%%%%%%%%%%%%%%%%%%%%%%%%%%%

%\begin{document}

%\maketitle

\begin{abstract}

Hierarchical reinforcement learning methods offer a powerful means of planning flexible behavior in complicated domains. However, learning an appropriate hierarchical decomposition of a domain into subtasks remains a substantial challenge. We present a novel algorithm for subtask discovery, based on the recently introduced multitask linearly-solvable Markov decision process (MLMDP) framework. The MLMDP can perform never-before-seen tasks by representing them as a linear combination of a previously learned basis set of tasks. In this setting, the subtask discovery problem can naturally be posed as finding an optimal low-rank approximation of the set of tasks the agent will face in a domain. We use non-negative matrix factorization to discover this minimal basis set of tasks, and show that the technique learns intuitive decompositions in a variety of domains. Our method has several qualitatively desirable features: it is not limited to learning subtasks with single goal states, instead learning distributed patterns of preferred states; it learns qualitatively different hierarchical decompositions in the same domain depending on the ensemble of tasks the agent will face; and it may be straightforwardly iterated to obtain deeper hierarchical decompositions.

\end{abstract}

\section{Introduction}

Hierarchical reinforcement learning methods hold the promise of faster learning in complex state spaces and better transfer across tasks, by exploiting planning at multiple levels of detail \cite{Barto2003}. A taxi driver, for instance, ultimately must execute a policy in the space of torques and forces applied to the steering wheel and pedals, but planning directly at this low level is beset by the curse of dimensionality. Algorithms like HAMS, MAXQ, and the options framework permit powerful forms of hierarchical abstraction, such that the taxi driver can plan at a higher level, perhaps choosing which passengers to pick up or a sequence of locations to navigate to  \cite{Sutton1999,Dietterich2000,Parr1998}. While these algorithms can overcome the curse of dimensionality, they require the designer to specify the set of higher level actions or subtasks available to the agent. Choosing the right subtask structure can speed up learning and improve transfer across tasks, but choosing the wrong structure can slow learning \cite{Solway2014,Brunskill2014}. The choice of hierarchical subtasks is thus critical, and a variety of work has sought algorithms that can automatically discover appropriate subtasks.

One line of work has derived subtasks from properties of the agent's state space, attempting to identify states that the agent passes through frequently \cite{Stolle2002a}. Subtasks are then created to reach these bottleneck states  \cite{VanDijk2011a,Solway2014,Diuk2013}. In a domain of rooms, this style of analysis would typically identify doorways as the critical access points that individual skills should aim to reach \cite{Simsek2009}. This technique can rely only on passive exploration of the agent, yielding subtasks that do not depend on the set of tasks to be performed, or it can be applied to an agent as it learns about a particular ensemble of tasks, thereby suiting the learned options to a particular task set.

Another line of work converts the target MDP into a state transition graph. Graph clustering techniques can then identify connected regions, and subtasks can be placed at the borders between connected regions \cite{Mannor2004}. In a rooms domain, these connected regions might correspond to rooms, with their borders again picking out doorways. Alternately, subtask states can be identified by their betweenness, counting the number of shortest paths that pass through each specific node \cite{Simsek2009,Solway2014}. Finally, other methods have grounded subtask discovery in the information each state reveals about the eventual goal \cite{VanDijk2011a}. Most of these approaches aim to learn options with a single or low number of termination states, can require high computational expense \cite{Solway2014}, and have not been widely used to generate multiple levels of hierarchy (but see \cite{Vigorito}).

Here we describe a novel subtask discovery algorithm based on the recently introduced Multitask linearly-solvable Markov decision process (MLMDP) framework \cite{Saxe2016}, which learns a basis set of tasks that may be linearly combined to solve tasks that lie in the span of the basis \cite{Todorov2009}. We show that an appropriate basis can naturally be found through non-negative matrix factorization \cite{Lee1999,Lee2000}, yielding intuitive decompositions in a variety of domains. Moreover, we show how the technique may be iterated to learn deeper hierarchies of subtasks.

\section{Background: The Multitask LMDP}
In the multitask framework of \cite{Saxe2016}, the agent faces a set of tasks where each task has an identical transition structure, but different terminal rewards, modeling the setting where an agent pursues different goals in the same fixed environment. Each task is modeled as a finite-exit LMDP \cite{Todorov2009}. The LMDP is an alternative formulation of the standard MDP that carefully structures the problem formulation such that the Bellman optimality equation becomes linear in the exponentiated cost-to-go. As a result of this linearity, optimal policies compose naturally: solutions for rewards corresponding to linear combinations of two optimal policies are simply the linear combination of their respective desirability functions \cite{Todorov2009bb}. This special property of LMDPs is exploited by \cite{Saxe2016} to develop a multitask reinforcement learning method that uses a library of basis tasks, defined by their boundary rewards, to perform a potentially infinite variety of other tasks--any tasks that lie in the subspace spanned by the basis can be performed optimally. 

Briefly, the LMDP \cite{Todorov2009,Todorov2009bb} is defined by a three-tuple $L=\langle S, P, R\rangle$, where $S$ is a set of states, $P$ is a passive transition probability distribution $P: S \times S \rightarrow [0,1]$, and $R$ is an expected instantaneous reward function $R: S \rightarrow \mathbb R$. The `action' chosen by the agent is a full transition probability distribution over next states, $a(\cdot|s)$. A control cost is associated with this choice such that a preference for energy-efficient actions is inherently specified. Finally, the LMDP has rewards $r_i(s)$ for each interior state, and $r_b(s)$ for each boundary state in the finite exit formulation. The LMDP can be solved by finding the \textit{desirability} function $z(s)=e^{V(s)/\lambda}$ which is the exponentiated cost-to-go function for a specific state $s$. Here $\lambda$ is a parameter related to the stochasticity of the solution. Given $z(s)$, the optimal control can be computed in closed form (see \cite{Todorov2006} for details). Despite the restrictions inherent in the formulation, the LMDP is generally applicable; see the supplementary material in \cite{Saxe2016} for examples of how the LMDP can be applied to non-navigational and conceptual tasks. A primary difficulty in translating standard MDPs into LMDPs is the construction of the action-free passive dynamics $P$; however, in many cases, this can simply be taken as the resulting Markov chain under a uniformly random policy.

The Multitask LMDP \cite{Saxe2016} operates by learning a set of $N_t$ tasks, defined by LMDPs $L_t=\langle S, P, q_i, q_b^t\rangle$, $t=1,\cdots,N_t$ with identical state space, passive dynamics, and internal rewards, but different instantaneous exponentiated boundary reward structures $q^t_b=\exp(r_b^T/\lambda), ~ t = 1, \cdots, N_t$. The set of LMDPs represent an ensemble of tasks with different ultimate goals. We can define the task basis matrix $Q=\left[ q^1_b ~ q^2_b ~ \cdots ~ q^{N_t}_b \right]$ consisting of the different exponentiated boundary rewards. Solving these LMDPs gives a set of desirability functions $z^t_i,~t=1,\cdots,N_t$ for each task, which can be formed into a desirability basis matrix $Z = \left[z^1_i~z^2_i~\cdots~z^{N_t}_i\right]$ for the multitask module. With this machinery in place, if a new task with boundary reward $q$ can be approximately expressed as a linear combination of previously learned tasks, $q \approx  Qw$. Then the same weighting can be applied to derive the corresponding optimal desirability function, $z = Zw$, due to the compositionality of the LMDP.

\subsection{Stacking the MLMDP}
\label{subsection_stacking}
The multitask module can be stacked to form deep hierarchies \cite{Saxe2016} by iteratively constructing higher order MLMDPs in which higher levels select the instantaneous reward structure that defines the current task for lower levels in a feudal-like architecture. This recursive procedure is carried out by firstly augmenting the layer $l$ state space $\tilde S^l=S^l \cup S_t^l$ with a set of $N_t$ terminal boundary states $S_t^l$ called \textit{subtask} states. Transitioning into a subtask state corresponds to a decision by the layer $l$ MLMDP to access the next level of the hierarchy. These subtask transitions are governed by a new $N^{l}_t$-by-$N^{l}_i$ passive dynamics matrix $P^{l}_t$. In the augmented MLMDP, the full passive dynamics are taken to be $\tilde{P^l}=[P_i^l;P_b^l;P_t^l]$, corresponding to transitions to interior states, boundary states, and subtask states respectively. Higher layer transitions dynamics $[P_i^{l+1};P_b^{l+1}]$ are then suitably defined \cite{Saxe2016}. Crucially, in order to stack these modules, both the subtask states themselves $S_t^l$, and the passive dynamic matrix $P_t^l$ must be defined. These are typically hand crafted at each level. 

\begin{figure*}[!ht]
\begin{center}
\includegraphics[width=\textwidth]{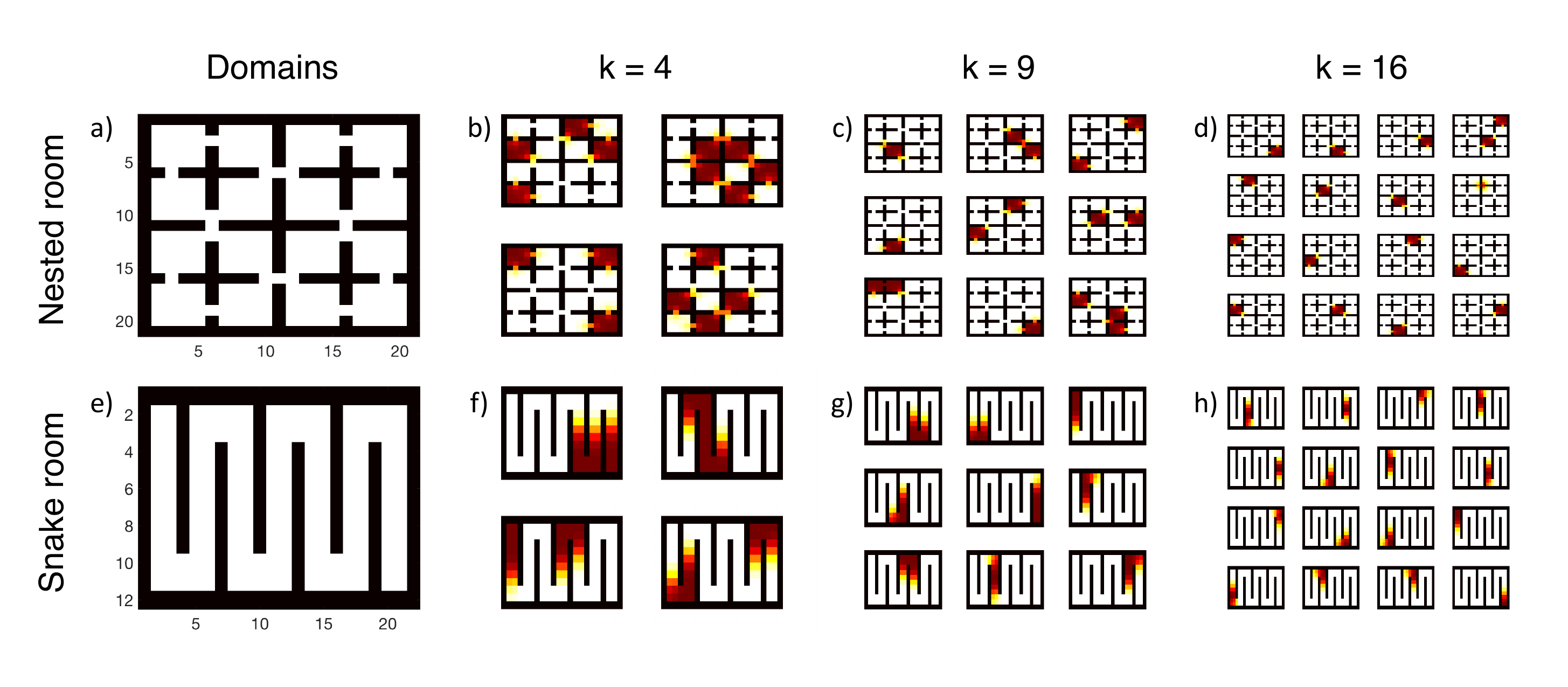}
\end{center}
\caption{ Intuitive decompositions in structured domains. b) c) d) Representations of the subtasks uncovered for decomposition factors $k=[4,9,16]$ in the nested-rooms domain respectively. Subtasks correspond to regions rather than single goal states, and typically find `rooms' rather than `doorways'. f) g) h) Representations of the subtasks uncovered in the snake-rooms domain. In all cases the subtasks uncovered are refactored for different values of $k$, and considered as a whole provide an approximate cover for the full action space.}
\label{fig_basic_panel}
\end{figure*}

\section{Subtask discovery via non-negative matrix factorization}
\label{sec_non_negative_matrix_factorization}

\begin{figure*}[ht]
\begin{center}
\includegraphics[width=\textwidth]{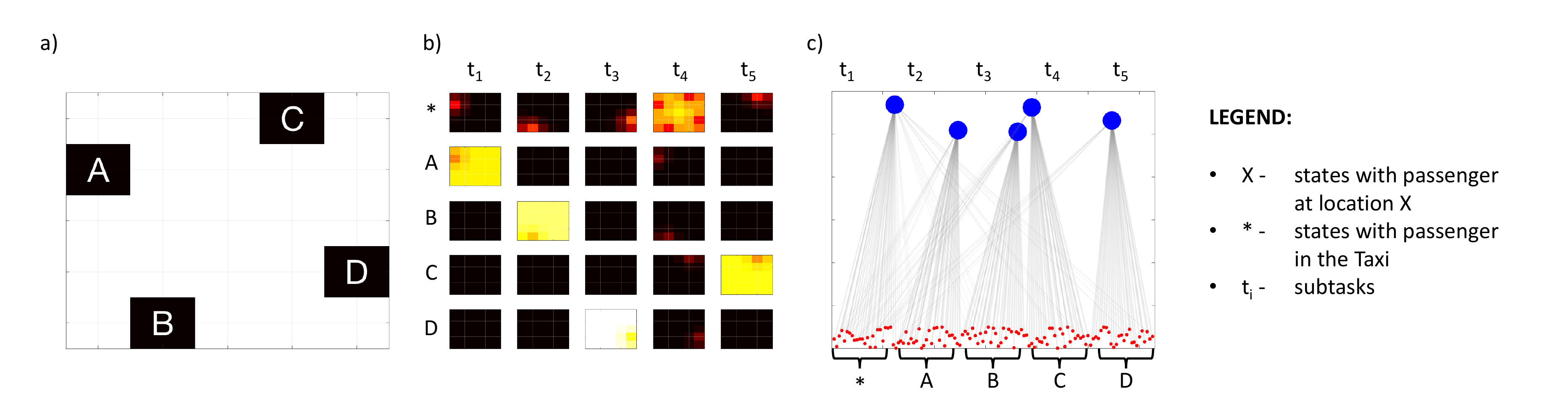}
\end{center}
\caption{ Decomposition in the taxi domain for $k=5$. a) The taxi domain with pick-up/drop-off locations labeled. b) A view of subtasks $t_1,\dots t_5$ divided into five copies of the base domain defined by the passenger's location. Subtasks correspond to all states in which the passenger is at a given location (regardless of the taxi's location), or in the taxi. c) A graphical representation of the subtask structure in which edge thickness is proportional to $D_{t_i}(s)$. Probability leakage into the pick-up/drop-off actions is clear. }
\label{fig_taxi_panel}
\end{figure*}

Prior work has assumed that the task basis $Q$ is given \textit{a priori} by the designer. Here we address the question of how a suitable basis may be learned. A natural starting point is to find a basis that retains as much information as possible about the ensemble of tasks to be performed, analogously to how principal component analysis yields a basis that maximally preserves information about an ensemble of vectors. In particular, to perform new tasks well, the desirability function for a new task must be representable as a (positive) linear combination of the desirability basis matrix $Z$. This naturally suggests decomposing $Z$ using PCA (i.e., the SVD) to obtain a low-rank approximation that retains as much variance as possible in $Z$. However, there is one important caveat: the desirability function is the exponentiated cost-to-go, such that $Z=\exp(V/\lambda)$. Therefore $Z$ must be non-negative, otherwise it does not correspond to a well-defined cost-to-go function.

Our approach to subtask discovery is thus to uncover a low-rank representation through non-negative matrix factorization, to realize this positivity constraint \cite{Lee1999,Lee2000}. We seek a decomposition of $Z$ into a data matrix $D \in \mathbf{R}^{(m \times k)}$ and a weight matrix $W \in \mathbf{R}^{(k \times n)}$ as:
\beq
	Z \approx DW
\eeq
where $d_{ij},w_{ij}\geq 0$. The value of $k$ in the decomposition must be chosen by a designer to yield the desired degree of abstraction, and is referred to as the \textit{decomposition factor}. Since $Z$ may be strictly positive, the non-negative decomposition is not unique for any $k$ \cite{Donoho2004}. Formally then we seek a decomposition which minimizes the cost function
\beq
	d_{\beta}(Z||DW),
    \label{eqn_beta_divergence}
\eeq

where $d$ is the $\beta$-divergence, a subclass of the more familiar Bregman Divergences \cite{Hennequin2011}. The $\beta$-divergence collapses to the better known statistical distances for $\beta \in \{ 0, 1, 2 \}$ corresponding to distances $\{ \text{`Itakura-Saito'}, \text{`Kullback-Leibler'},\text{`Euclidean'}\}$ \cite{Cichocki2011a}.
%This setup is also closely related to common clustering techniques. By including an additional orthogonality constraint on the weight matrix $WW^T=I$, the well-known k-means clustering procedure is recovered. Similarly, when the $\beta$-divergence is set to the Kullback-Leibler distance, probabilistic latent semantic analysis is recovered as the clustering procedure \cite{Ding2008a}.

Crucially, since $Z$ depends on the set of tasks that the agent will perform in the environment, the representation is defined by the tasks taken against it, and is not simply a factorization of the domain structure. 

% Through an appropriate scaling transformation of the resulting data matrix $D$, we can guarantee that the columns of $D$ may be realized as continuous actions in the LMDP framework; $Z\approx (D\Lambda)(\Lambda^{-1}W), \Lambda = \text{diag}\;(\sum_i d_{ij})$. As such we may interpret the columns of the scaled $D$ matrix as a generalized action basis, whose basis vectors may or may not corresponds to optimal actions for some subtask. This is a more general notion of the action basis than that considered in [cite: us], however the modular compositionality described therein remains applicable in this setting.

\subsection{Conceptual demonstration}

To demonstrate that the proposed scheme recovers an intuitive decomposition, we consider the resulting low-rank approximation to the action basis in two domains for a few decomposition factors. All results presented in this section correspond to solutions to Eqn.(\ref{eqn_beta_divergence}) for $\beta = 1$. In the same way that the columns of $Z$ correspond directly to the optimal actions for the subtasks defined in the task basis $Q$ \cite{Todorov2006}, so the columns of the low-rank approximation $D$ may be considered the {\textit{ generalized actions}} of the uncovered subtasks. In Fig. \ref{fig_basic_panel}, the subtasks uncovered for decomposition factors $k= \{4,9,16\}$ are overlaid onto the corresponding domain(s).

It is  clear from Figs. (\ref{fig_basic_panel}.b,c,d) that  the subtasks uncovered  in our scheme  correspond to  `rooms' rather than the perhaps more familiar `door-ways' uncovered by other schemes. This  is due to the  fact that our scheme can, and typically does, uncover more complex distribution patterns over states rather than isolated goal states. There are a few notable features of the decomposition. Since an approximate basis is uncovered, there is no implicit preference or value ordering to the subtasks - all that is important is that they provide a good subspace for the task ensemble. An associated fact is that the resulting decomposition is `refactored' for higher-rank decompositions; that is to say that $D_{k+1} \neq [D_k, d_{k+1}]$. 
%This is intuitively pleasing; as in humans new experiences are not {\emph{added}} to our knowledge of the world, but rather {\emph{integrated}} into our existing knowledge \cite{}.  

With some of the conceptual features of the scheme firmed up, we consider its application to the standard TAXI domain with one passenger and four pick-up/drop-off locations. The $5\times 5$ TAXI domain considered is depicted in Fig.(\ref{fig_taxi_panel}.a). Here the agent operates in the product space of the base domain ($5 \times 5 = 25$), and the possible passenger locations ($5$ choose $1 = 5$) for a complete state-space of $125$ states. We consider a decomposition with factor $k=5$. Figs.(\ref{fig_taxi_panel}.b,c) are complementary depictions of the same subtask structure uncovered by our scheme.

% \begin{figure*}[ht]
% \begin{center}
% \includegraphics[width=\textwidth]{fig_taxi_panel_v2.pdf}
% \end{center}
% \caption{ Decomposition in the taxi domain for $k=5$. [1] The taxi domain with pick-up/drop-off locations labeled. [2] A view of subtasks $t_1,\dots t_5$ divided into five copies of the base domain defined by the passenger's location. Subtasks correspond to all states in which the passenger is at a given location (regardless of the taxi's location), or in the taxi. [3] A graphical representation of the subtask structure in which edge thickness is proportional to $D_{t_i}(s)$. Probability leakage into the pick-up/drop-off actions is clear. }
% \label{fig_taxi_panel}
% \end{figure*}

The columns of Fig.(\ref{fig_taxi_panel}.b) are the subtasks visually divided into the five copies of the base domain defined by the passenger's location (here location $*$ corresponds to the passenger being in the taxi). Consideration of subtask $t_1$ shows that the generalized action corresponds to the region of all base states with the passenger at location $A$. 
%Again this is in stark contrast with the subtasks uncovered by other schemes in which the subtask would correspond to getting the taxi to location A, regardless of the passenger's location \cite{}. 
We also note the probability leakage into $(*,t_1)$ corresponding to the `pick-up' action. A similar analysis holds for the other subtasks. Considered as a whole, the subtask basis represents policies for getting the passenger to each of the pick-up/drop-off locations, and for having the passenger in the taxi.

Concretely then, the task of picking up the agent at location $A$, and transporting them to location $B$ would be realized by firstly weighting subtask $t_4$ (getting the passenger into the taxi) and then subtask $t_2$ (getting the passenger to location $B$, regardless of taxi location).  Worthy of special mention is subtask $t_4$ which corresponds to the passenger being in the taxi. Here the generalized action focuses probability mass at the center of the room, with a symmetric fall-off (corresponding to the symmetric placement of the pick-up/drop-off locations); again we note the probability leakage into the drop-off actions. Fig.(\ref{fig_taxi_panel}.c) depicts the same subtask structure graphically. Red balls depict states, while blue balls depict subtasks. The edge thickness is proportional to $D_{t_i}(s)$. The primary edge connections correspond to the `regions' identified in Fig.(\ref{fig_taxi_panel}.b); all base states in which the passenger is at a given location. Here the probability leakage is perhaps more apparent; observe how all subtask states in which the passenger is at a location map to states in which the passenger is in the taxi (corresponding to the pick-up action), and similarly the subtask in which the passenger is in the taxi maps to states in which the passenger is at each location (corresponding to the drop-off action).

\section{Hierarchical decompositions}
\label{sec_hierarchical_decompositions}

The proposed scheme uncovers a set of subtasks by finding a low rank approximation to the desirability matrix Z. This procedure can simply be reapplied to find an approximate basis for each subsequent layer of the hierarchy, by factoring $Z^{l+1}$. However, as noted in section \ref{subsection_stacking}, in order to define $Z^{l+1}$ in the first place, both the subtasks $S_t^{l}$, and the subtask passive dynamics $P_t^{l}$ must be specified. 
%The decomposition in section \ref{sec_non_negative_matrix_factorization} suggests that non-negative matrix factorization of the action basis uncovers intuitive subtasks at the base layer. We wish to extend the reach of the method by recursing the procedure to form a deep hierarchy. To do so requires that we define a notion of an action basis over the subtasks. This is done by first defining an {\emph{action module}} as a multitask LMDP (MLDMP), and then stacking these disconnected modules to arbitrary depth \cite{Saxe2016}. The MLMDP is recursively defined, at layer $l$, in terms of the subtask transition matrix $P^l_t$. Lower-level MLMDPs are abstracted to form higher-level MLMDPs by choosing a set of `subtask' states which can be accessed by the lower level (the subtask transition probabilities $P_t$ are typically defined by a designer). Lower levels access these subtask states to indicate completion of a subgoal and to request more information from higher levels; higher levels communicate new subtask state instantaneous rewards, and hence the concurrent task blend, to the lower levels.

The subtask states $S_t^l$ may be directly associated with the generalized actions defined by the columns of $D^l$. Where the columns of $D^l$ corresponds to the desirability functions for a set of approximate basis tasks; these approximate basis tasks are taken to be the subtask states. Furthermore, as noted in section \ref{subsection_stacking}, in the original formulation, $P_t^{l}$ is hand-crafted by a designer for each layer \cite{Saxe2016}. Here we relax that requirement and simply define the subtask transitions as
\beq
	P_t^{l} = \alpha^{l} D^{l},
    \label{eqn_subtask_transition_probability}
\eeq
where $\alpha^l$ is a hand-crafted scaling parameter which controls how frequently the agent will transition to the higher layer(s).

\begin{figure}[ht!]
\begin{center}
\includegraphics[width=0.45\textwidth]{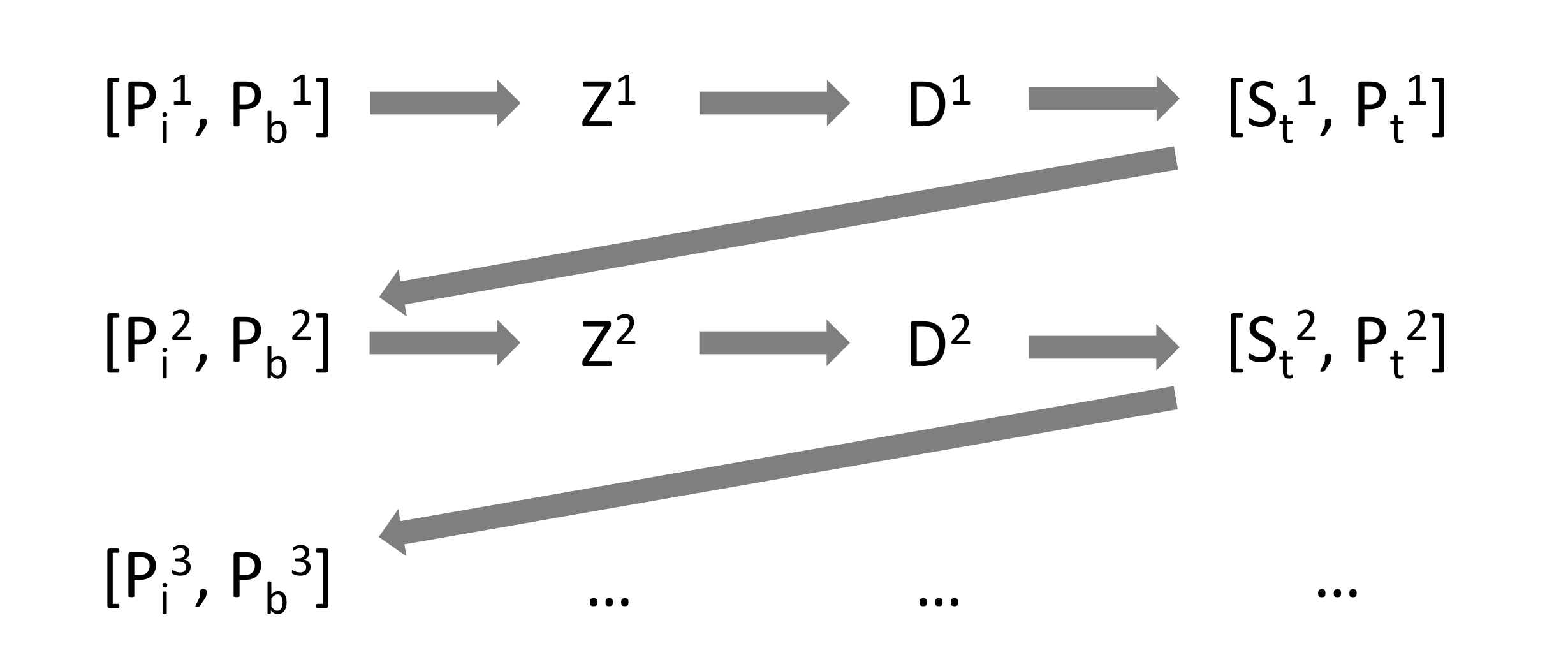}
\end{center}
\caption{ A recursive procedure for constructing hierarchical subtasks. By associating the subtask states $S_t^l$ of the MLMDP with the generalized actions corresponding to columns of $D^l$, the subtask discovery mechanism may be recursed to uncover hierarchical subtasks corresponding to ever greater levels of abstraction.}
\label{hierarchical_flow}
\end{figure}

\begin{figure}[!ht]
\begin{center}
\includegraphics[width=0.45\textwidth]{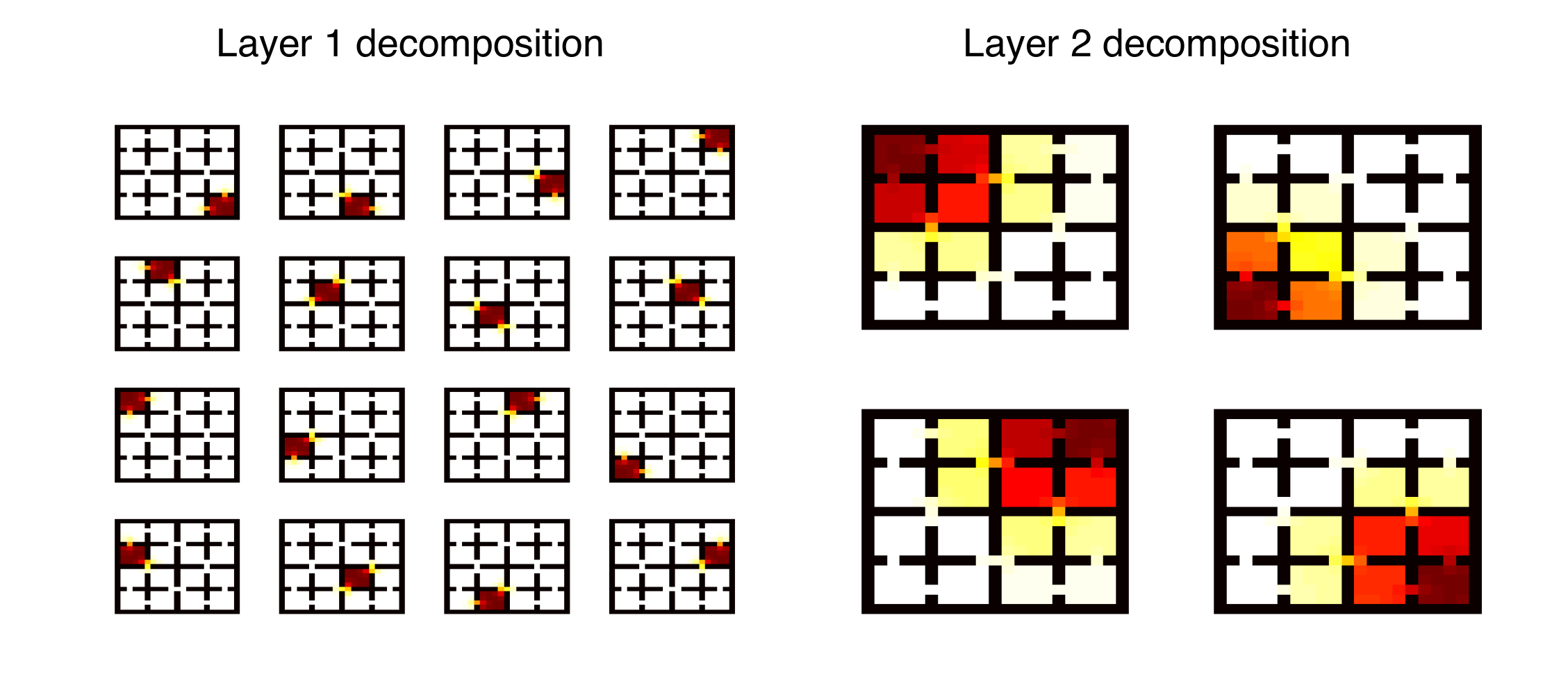}
\end{center}
\caption{ Hierarchical decomposition of the nested rooms domain. Recursive application of the scheme yields intuitive results. The first layer of abstraction uncovers `rooms'; the second layer uncovers quadrants. }
\label{fig_nested_rooms_hierarchy}
\end{figure}

A powerful intuitive demonstration of the recursive potential of the scheme is had by considering firstly the $k=16$ decomposition of the nested rooms domain, Fig.(\ref{fig_basic_panel}), followed by the $k=4$ decomposition of the higher layer desirability matrix, computed by solving the higher layer MLMDP. The decomposition at the first layer intuitively uncovers a subtask for each of the sixteen rooms in the domain, Fig.(\ref{fig_nested_rooms_hierarchy}); the decompositions of the second layer uncovers the abstracted quadrants. As such planning at the highest layer will drive the agent to the correct quadrant, whereas planning at the lower layer will drive the agent to a specific room, and planning at the level of primitives will then navigate the agent to the specific state.

\begin{figure}[ht]
\begin{center}
\includegraphics[width=0.45\textwidth]{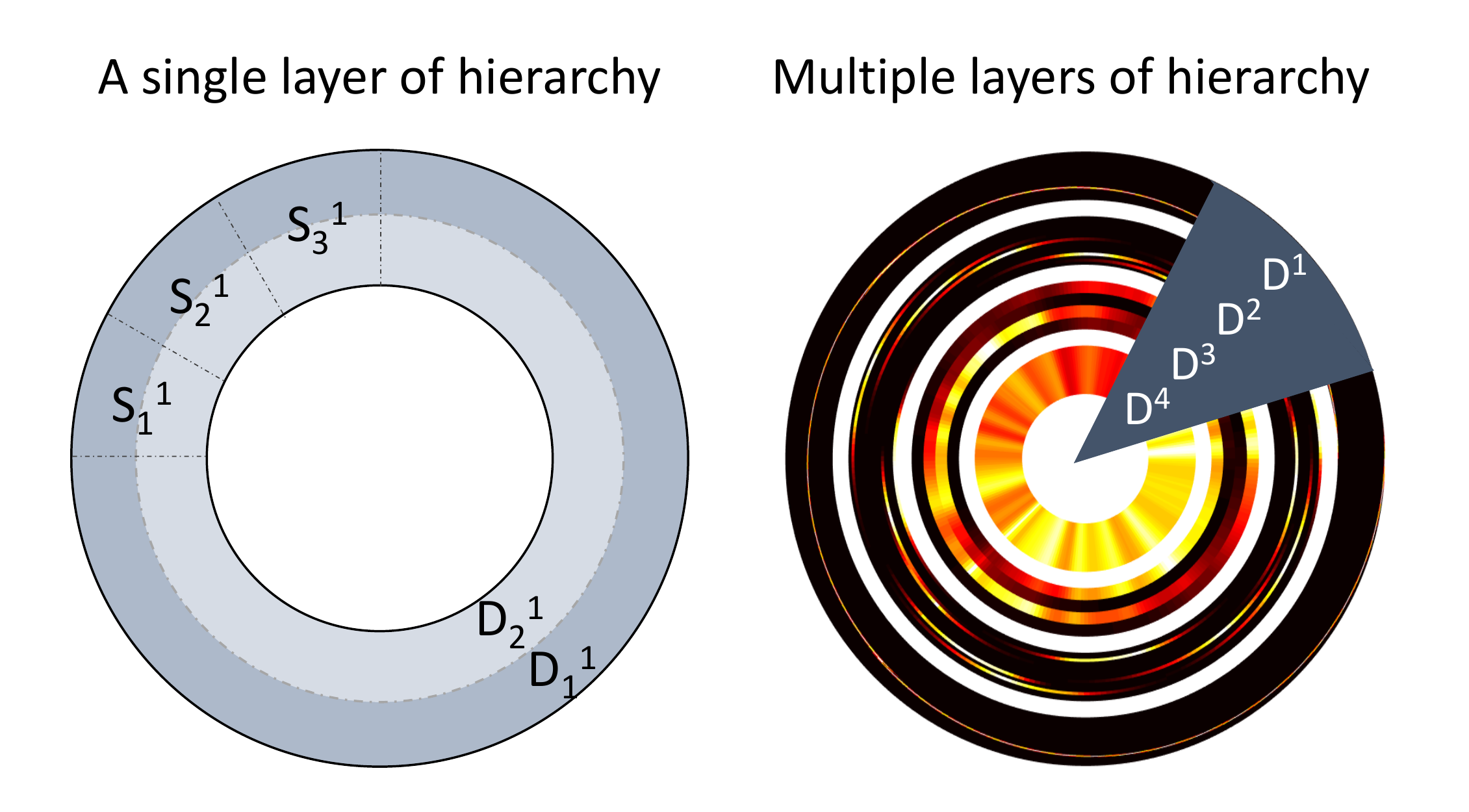}
\end{center}
\caption{ Deep hierarchical decomposition of the 1D ring domain. Each ring represents the full state space, onto which successively higher layer decompositions have been overlayed. The outer most ring corresponds to the first layer of abstraction; here subtasks are strongly localized indicating generalized actions that would drive the agent to specific fine-grained regions. Inner rings correspond to subsequent layers of abstraction; here subtasks exhibit more distributed behaviour indicating generalized actions that would drive the agent to broader patches of the states space. }
\label{fig_1dring}
\end{figure}

To show that the scheme uncovers sensible decompositions when applied to deeper hierarchies, we consider a 1D ring of 256 states in Fig.(\ref{fig_1dring}). At each layer $l$, we perform the decomposition with factor $k^l=\frac{256}{4^{(4-l)}}$. The subtasks uncovered in $D^l$ are then overlayed onto the base domain. At lower levels of abstraction (outer rings), subtasks exhibit strong localized behaviour; whereas at higher levels of abstraction (inner rings), the subtasks uncovered correspond to broad, complex distributions over states, covering whole regions of the state space. 

\section{Determining the decomposition factor $k$}
\label{subsection_decomposition_factor}

Further leveraging the unique construction in Eqn.(\ref{eqn_beta_divergence}) we may formally determine the optimal decomposition factor $k$ by critiquing the incremental value of ever higher rank approximations to the complete action basis. Let us denote the dependence of $d_{\beta}(\cdot)$ on the decomposition factor simply as $f(k)$. Then we may naively define the optimal value for $k$ as the smallest value that demonstrates diminishing incremental returns through classic elbow-joint behaviour
\beq
	|f(k+1)-f(k)| < |f(k)-f(k-1)|.
    \label{eqn_optimal_decomposition}
\eeq
In practice, when the task ensemble is drawn uniformly from the domain, the observed elbow-joint behaviour is an encoding of the high-level domain structure.

\begin{figure}[ht]
\begin{center}
\includegraphics[height=4cm,width=0.3\textwidth]{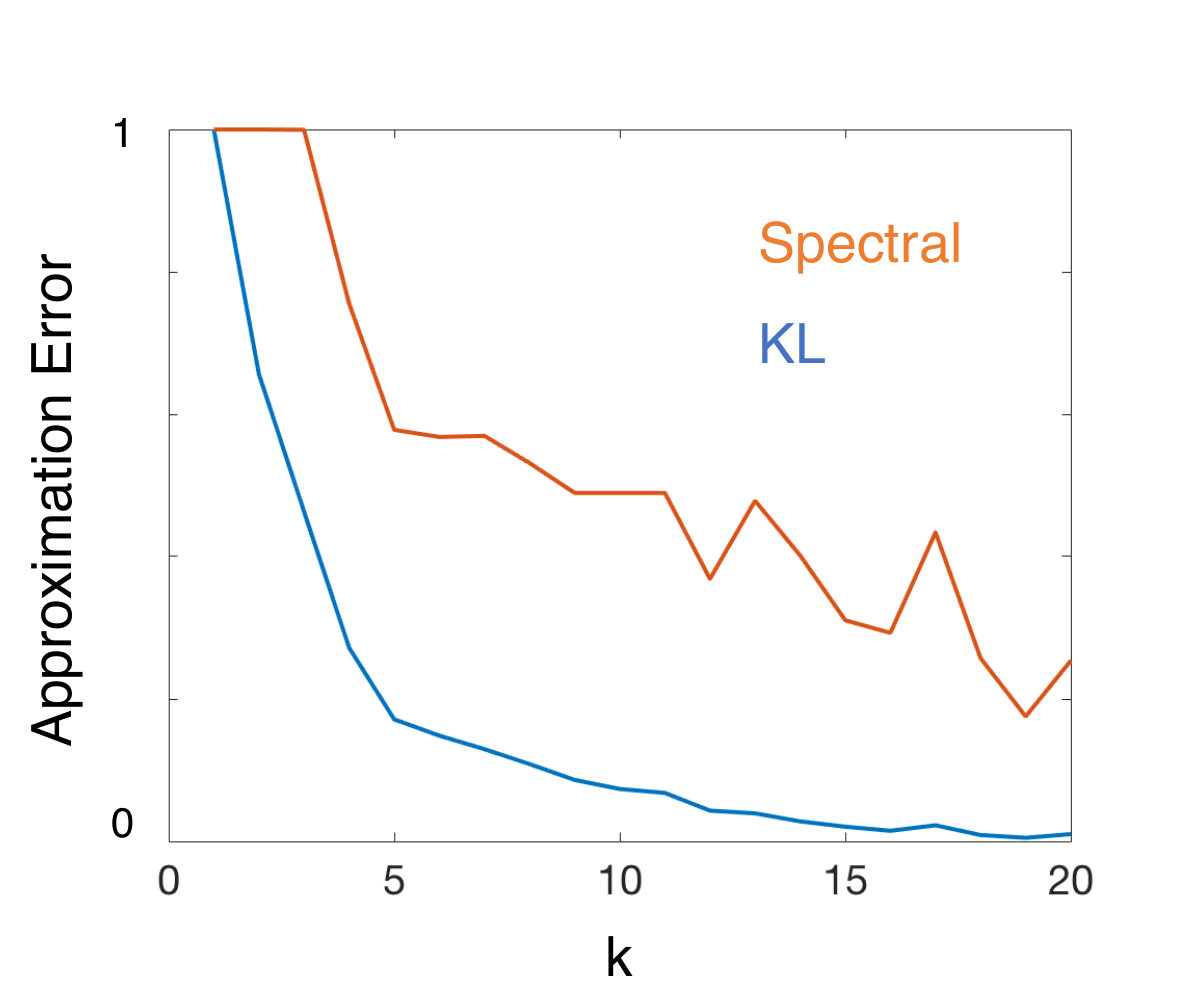}
\end{center}
\caption{ Normalized approximation error for higher rank decompositions in the TAXI domain contains task and domain information. Notice the `elbow-joint' behaviour at $k=5$: this indicates diminishing incremental returns for higher rank approximations. This is intuitive since five is the minimum number of DOF required to capture the dynamics for each passenger-location configuration. }
\label{fig_rank_error}
\end{figure}

The normalized approximation error for higher rank approximations in the TAXI domain is considered in Fig.(\ref{fig_rank_error}). Both measures exhibit `elbow-joint' behaviour at $k=5$. This result is intuitive; we would expect to see a subtask corresponding to the pick-up action in the base MDP (this being the state of having the passenger in the taxi), and a subtask corresponding to the drop-off action in the base MDP (this being the state of having the passenger at each location). This critical value would be identified by Eqn.(\ref{eqn_optimal_decomposition}).

\section{Equivalence of subtasks}

The new paradigm allows for a natural notion of subtask equivalence. Suppose some standard metric is defined on the space of matrices in $\mathbf{R}^{m \times n}$ as $m(A,B) = ||A-B||_2^2$. Then a formal pseudo-equivalence relation may be defined on the set of subtasks, encoded as the scaled columns of the data matrix $D$, as $A\sim B \rightarrow m(A,B) < \epsilon$. The pseudo-equivalence class follows as
\beq
	\{ (D_j,W_j) \in d_{\beta}(Z,D_jW_j) \;\; | \;\; (D_iW_i) \sim (D_jW_j)  \}.
\eeq
This natural equivalence measure allows for the explicit comparison of different sets of subtasks.

\begin{figure}[ht]
\begin{center}
\includegraphics[height=4cm, width=0.3\textwidth]{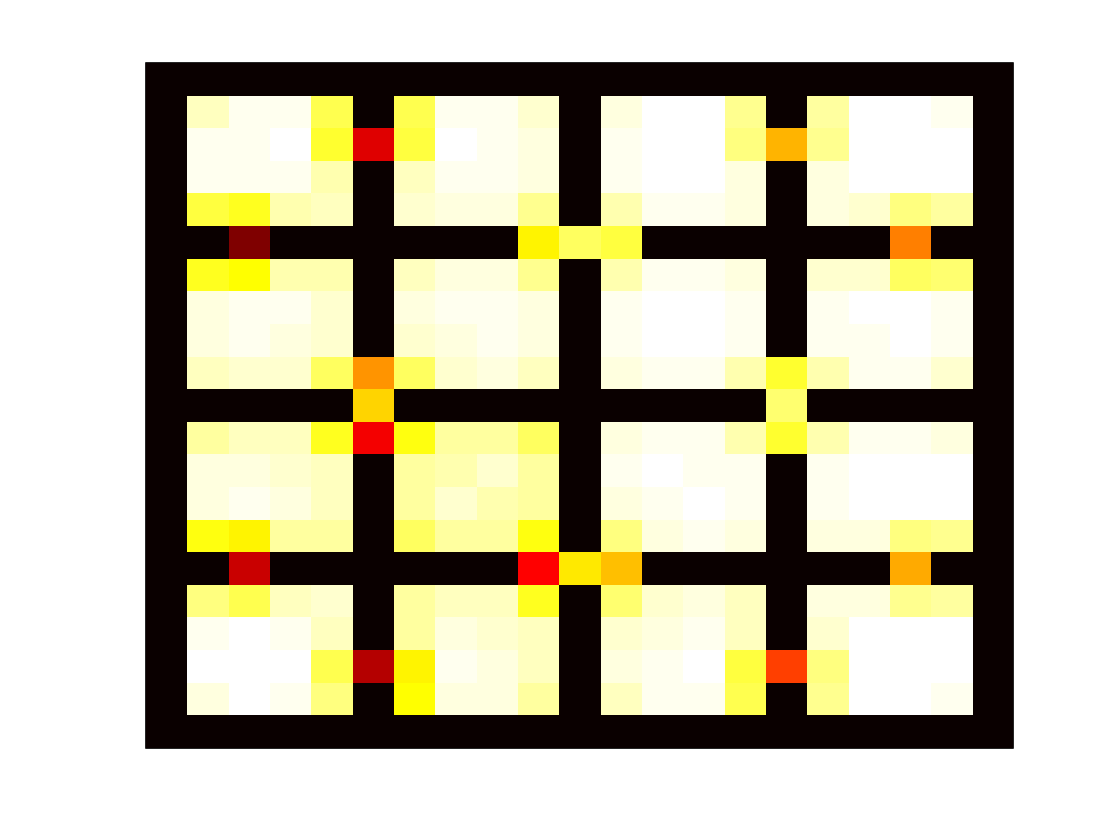}
\end{center}
\caption{ A natural duality exists between our decomposition and `doorways'. Consideration of representation switching between states uncovers `doorways' as boundary states of our approximate basis.  }
\label{fig_nested_rooms_doorways}
\end{figure}

As noted above, our scheme uncovered `rooms', where other methods typically uncover `doorways', see Fig.(\ref{fig_basic_panel}). There is a natural duality between these abstractions. By considering the states whose representation in Eqn.(\ref{eqn_beta_divergence}), $w_s$, changes starkly on transitions we uncover those states which constitute the boundary between similar `regions'.  Explicitly we consider the function $g:S\rightarrow \mathbf{R}$:
\beq
	g(s) = \sum_i p_{is}||w_i -w_s||_2^2,
\eeq
which is a weighted measure of how the representations of neighbour states differ from  the current state. States for which $g(s)$ takes a high value are those on the boundary between `regions'. A cursory analysis of Fig.(\ref{fig_nested_rooms_doorways}) immediately identifies doorways as being those boundary states.

\section{Conclusion}
We present a novel subtask discovery mechanism based on the low rank approximation of the action basis afforded by the LMDP framework. The new scheme reliably uncovers intuitively pleasing decompositions in a variety of sample domains. The proposed scheme is fundamentally dependent on the task ensemble, and may be straightforwardly iterated to yield hierarchical abstractions. Moreover the unusual construction allows us to analytically probe a number of natural questions inaccessible to other methods; we consider specifically a measure of the equivalence of different set of subtasks, and a quantitative measure of the incremental value of greater abstraction.

%{\bf Acknowledgements} NA

\small

%\bibliography{Mendeley_Subtask_Discovery}
%\bibliographystyle{unsrt}

%%%%%
%My bibliography hack for arxiv
%%%%%

%%%%

\bibliographystyle{icml2017}

\end{document}